\documentclass[12pt,a4paper]{article}

\usepackage{enumitem}

\usepackage[T1]{fontenc} 
\usepackage{tgtermes} 

\usepackage{tabularray} 
\usepackage{graphicx} 
\usepackage[margin=25mm]{geometry}
\usepackage[hyphens]{url}
\usepackage[hidelinks]{hyperref}
\usepackage[hyphenbreaks]{breakurl}

\usepackage{amsmath}
\usepackage{amsfonts}
\usepackage{amssymb}
\usepackage{siunitx}
\usepackage{verbatim}
\usepackage{natbib} 

\title{Taking control: \\
Policies to address extinction risks from advanced AI}
\author{
  Andrea Miotti\\
  \small{Control AI \& Conjecture}\\
  \small\texttt{andrea@controlai.com}}

\begin{document}
\maketitle

\begin{abstract}

\noindent
This paper provides policy recommendations to reduce extinction risks from advanced artificial intelligence (AI). First, we briefly provide background information about extinction risks from AI. Second, we argue that voluntary commitments from AI companies would be an inappropriate and insufficient response. Third, we describe three policy proposals that would meaningfully address the threats from advanced AI: (1) establishing a Multinational AGI Consortium to enable democratic oversight of advanced AI (“MAGIC”), (2) implementing a global cap on the amount of computing power used to train an AI system (“global compute cap”), and (3) requiring \textit{affirmative safety evaluations} to ensure that risks are kept below acceptable levels (“gating critical experiments”). MAGIC would be a secure, safety-focused, internationally-governed institution responsible for reducing risks from advanced AI and performing research to safely harness the benefits of AI. MAGIC would also maintain emergency response infrastructure (“kill switch”) to swiftly halt AI development or withdraw model deployment in the event of an AI-related emergency. The global compute cap would end the corporate race toward dangerous AI systems while enabling the vast majority of AI innovation to continue unimpeded. Gating critical experiments would ensure that companies developing powerful AI systems are required to present affirmative evidence that these models keep extinction risks below an acceptable threshold. After describing these recommendations, we propose intermediate steps that the international community could take to implement these proposals and lay the groundwork for international coordination around advanced AI.

\end{abstract}
\newpage
\section{Executive summary}

\textbf{Advanced AI poses an extinction risk to humanity}. Leading AI researchers and CEOs of the three most advanced AI companies have recognized these risks. The UK AI Safety Summit provides world leaders with an opportunity to lay the groundwork for international coordination around these pressing threats. 

\textbf{Extinction Risks from AI arise chiefly from “scaling”}. Scaling refers to the process of building increasingly large, ever more autonomous, and more opaque AI systems. The vast majority of AI research with concrete applications, such as cancer detection and trading algorithms, is unconcerned with scaling. 

\textbf{Governments have a critical and time-limited opportunity to reduce extinction risks from AI}.  This report recommends immediate actions that can be taken by governments: recognizing the extinction risks from advanced AI, acknowledging the need for concrete scaling limits, and committing to build an international advanced AI project. Specifically, we recommend:

\begin{enumerate}
    \item \textbf{Establishing a Multinational AGI Consortium (MAGIC), a ‘CERN for AI’}. This institution would house world-leading AI scientists from all signatory countries, dedicated to the joint mission of working on advanced AI safety. MAGIC would perform cutting-edge research to control powerful AI systems and establish emergency response infrastructure that allows world leaders to swiftly halt AI development or deployment. 
    \item \textbf{Implementing global compute limitations}. Compute Limitations mitigate extinction risk by throttling the very few dangerous AI systems that rely on advanced hardware, while leaving the vast majority of the AI ecosystem unhindered. We recommend a tiered approach to compute limitations: development above a \textbf{moratorium threshold} would not be allowed to occur (except within MAGIC), development above a \textbf{danger threshold} would need to be regulated by MAGIC or a MAGIC-certified regulator, and development below the danger threshold would be unaffected by MAGIC. 
    \item \textbf{Gating critical experiments}. Before developing critical systems, companies should demonstrate the safety of these systems through verifiable criteria. The burden of proof should be on companies performing Frontier AI research to demonstrate that these systems are safe (rather than on regulators or auditors to show that systems are dangerous). This follows the standard practices from high-risk sectors, where demonstrating safety is a precondition for undertaking high-risk endeavors.

\end{enumerate}

These proposals have strong support among the British and American public. Recent YouGov polling shows that 83\% believe AI could accidentally cause a \textbf{catastrophic event}, 82\% prefer \textbf{slowing down the development of AI,} 82\% \textbf{do not trust AI tech executives to regulate AI}, 78\% support a \textbf{“global watchdog”} to regulate powerful AI, and 74\% believe AI policy should currently \textbf{prevent AI from “reaching superhuman capabilities”} (Artificial Intelligence Policy Institute [AIPI], 2023; Control AI, 2023).

\newpage
\section{Introduction}\label{sec:intro}
\subsection{Extinction risks from advanced AI} 
\textbf{Advanced AI poses an extinction risk to humanity.} Leading AI researchers, alongside the CEOs of the three main advanced AI companies, all have recently signed a statement acknowledging: \textbf{“Mitigating the risk of extinction from AI should be a global priority alongside other societal-scale risks such as pandemics and nuclear war”} (Center for AI Safety, 2023). Sam Altman, CEO of OpenAI, stated that the bad case from AI is “lights out for all of us” (Altman, 2023). Dario Amodei, CEO of Anthropic, claimed that the chance of a civilization-scale catastrophe was at around 10-25\% (The Logan Bartlett Show, 2023). Geoffrey Hinton, considered a godfather of modern AI, recently quit Google to warn about the extinction risks from AI (MIT Technology Review, 2023). 

\textbf{Extinction risks from AI arise from “scaling”.} Scaling refers to the process of building increasingly large, more autonomous, and more opaque AI systems. The vast majority of AI research with concrete applications, such as cancer detection and trading algorithms, is unconcerned with scaling. The extreme risks are specific to advanced AI (sometimes called “Frontier AI”), which focuses on scaling. Companies that focus on scaling are attempting to create AI that surpasses human performance (sometimes called artificial general intelligence, AGI, human-level machine intelligence, artificial superintelligence, or superhuman AI.) Ian Hogarth, Chair of the UK’s AI Foundation Model Taskforce, more aptly described them as “godlike AI” due to the immense capabilities they would hold, and he noted that stopping the scaling race to godlike AI is a crucial policy priority (Hogarth, 2023).

\textbf{Extinction risks could occur very soon.} Turing Award winner Yoshua Bengio has stated that loss of control to rogue AI systems could emerge in “as little as a few years” unless appropriate precautions are taken. According to Anthropic CEO Dario Amodei, systems that enable many more actors to carry out large-scale biological attacks are likely to be built within two to three years (Oversight of A.I.: Principles for Regulation, 2023a; 2023b).

\textbf{World leaders have a fleeting window of opportunity to prevent extinction risks from AI. Risks from advanced AI} are widely acknowledged, and the public supports regulation (AIPI, 2023; Center for AI Safety, 2023 Control AI, 2023). Frontier AI systems are powerful enough to inspire action but not yet powerful enough to pose extinction risks, and only a few companies are capable of developing Frontier AI systems. These are favorable conditions for regulation. They may not last long, underscoring the need for swift and urgent action.

\textbf{To ensure global security, Frontier AI progress should occur in a secure international facility that conducts cutting-edge research while keeping extinction risks below an acceptable level.} The first step toward this future involves engaging in international coordination. 

\hfill\begin{minipage}{\dimexpr\textwidth-1cm}
“I will refer to it as what is: \textbf{God-like AI}. A superintelligent computer that learns and develops autonomously, that understands its environment without the need for supervision and that can transform the world around it. \textbf{God-like AI could be a force beyond our control or understanding, and one that could usher in the obsolescence or destruction of the human race.}” – Hogarth (2023)
\end{minipage}

\subsection{The UK AI Safety Summit: A Pivotal Moment }\label{sec:methods}
\hfill\begin{minipage}{\dimexpr\textwidth-1cm}
“\textbf{It felt deeply wrong that consequential decisions potentially affecting every life on Earth could be made by a small group of private companies without democratic oversight}. Did the people racing to build the first real AGI have a plan to slow down and let the rest of the world have a say in what they were doing?” – Hogarth (2023)
\end{minipage}
\hspace{1cm}

The Summit represents a historic opportunity to reduce extinction risks through national and international measures. The Summit’s focus on \textbf{Frontier AI systems} and \textbf{extinction risks} is crucial, given the nature of these threats and the need for urgent action. The Summit’s emphasis on international coordination underscores the great need for a coordinated global response to this major global security threat.

\textbf{This is an opportune moment to act}. Extinction risks from advanced AI are widely acknowledged by experts, the public is calling for regulation, and computing resources are still expensive and physically concentrated, constituting a natural bottleneck for government intervention. Furthermore, progress in AI is rapid and exponential: waiting until risks are imminent is not possible when dealing with technology that moves at an exponential pace. Notably, many experts believe extreme dangers could plausibly emerge within the next 2-5 years (e.g., Oversight of A.I.: Principles for Regulation, 2023a; Leike \& Sutskever, 2023). Taken together, \textbf{the present moment represents an opportune, unprecedented, and necessary moment to develop the national and international regulations needed to reduce extinction risks from advanced AI}. 

The Summit has the potential to usher in a series of concrete international agreements that could substantially reduce extinction risks from Frontier AI. 

There are three potential categories of outcomes for the AI Safety Summit:

\begin{itemize}
    \item \textbf{Insufficient: Extinction risk from AI fails to be addressed}: Voluntary measures that do not address extinction risks are endorsed. Governments fail to agree on common principles. The measures endorsed rely on companies to govern themselves, lack proper enforcement, and fail to place the burden of proof on companies to show that their activities are safe. Scaling is allowed to continue indefinitely until some unspecified point in the future, when we are much closer to imminent danger. 
    \item \textbf{\textbf{Adequate:} Extinction risk is acknowledged, and steps forward are agreed to in principle}: The AI Safety Summit attendees all agree to acknowledge that continued Frontier AI development poses an extinction risk to humanity. Given this, the Summit agrees in principle that rapid and effective measures must be taken by major governments to limit further AI scaling, that the safety of very advanced AI systems must be proven before they can be developed or deployed, and that a priority of the Summit going forward is to establish an international regime to deal with Frontier AI.  
   \item  \textbf{Excellent:} \textbf{Concrete measures are agreed upon to minimize extinction risk from AI}: The AI Safety Summit publicly acknowledges the risks from advanced AI. World leaders agree that there should be concrete scaling limits put in place and commit to ending the race to AGI. World leaders recognize the urgency of working on an international organization to regulate the development of advanced AI.  
\end{itemize}

\newpage
\section{Voluntary commitments and “responsible capabilities scaling” policies: solutions that do not address extinction risks}
\hfill\begin{minipage}{\dimexpr\textwidth-1cm}
“Responsible scaling” sounds like it came from the same lobbying firm that coined “clean coal.” – Tegmark (2023)
\end{minipage}
\hspace{1cm}

Responsible Scaling Policies (RSPs) are proposals suggesting voluntary commitments on internal protocols AI companies can use to manage risks. Broadly, RSPs involve using tests (dangerous capability evaluations) to determine how powerful and dangerous an AI system is. RSPs are sometimes referred to as “responsible capabilities scaling” or “risk-informed development policies”;for convenience, we will use the term RSP throughout the report, but our points also apply to voluntary scaling commitments that are branded under different labels. 

The RSP framework utilizes a burden of proof that is the opposite of what is a common standard in high-risk sectors.\textbf{ In RSPs, rather than it being the onus of the company to demonstrate that their plans for further scaling of powerful Frontier AI systems is safe, the onus is on auditors to demonstrate that they are dangerous.} The RSP system presumes that Frontier AI is safe until proven otherwise.  

So far, these have been proposed by Anthropic, one of the major AGI companies, and ARC Evaluations, an AI model testing organization (Anthropic, 2023; Alignment Research Center, 2023). RSPs have been briefly mentioned by a member of the UK government as a possible measure, and “responsible capabilities scaling” is listed among the discussion topics of the Summit.

In the context of extinction risks from Frontier AI, which are a core focus of the Summit and future government intervention, \textbf{RSPs are inadequate at addressing these risks.}

Below, we highlight a few of the properties that make RSPs (and other types of voluntary commitments) inadequate to address extinction risks from advanced AI.

\begin{enumerate}
    \item \textbf{RSPs reverse the standard burden of proof in high-risk sectors}. The RSP framework presumes that Frontier AI development is safe until proven otherwise. Under the RSP framework, AI development should be allowed to continue until auditors have clear evidence that models \textit{already possess} dangerous capabilities. This is a stark departure from norms in other areas of risk management, where we expect companies to provide \textit{affirmative evidence }of safety (see Raz \& Hillson, 2005 for a review of standard risk management practices). This is self-defeating in the context of extinction risks from Frontier AI: waiting until the risks are realized before intervening does not work in the case of extinction-level risks. Instead, common risk management practices demand evidence of \textit{affirmative safety} – a point elaborated on in the next section of this report. 
    \item \textbf{RSPs lack oversight, accountability, and enforcement}. The RSP framework presumes that companies should be responsible for managing extinction risks. The company is allowed to figure out if and how to manage the risks, and the company is allowed to decide what level of risk is worth incurring. There are no requirements for companies, no oversight to ensure that companies follow through on their voluntary commitments, and no accountability or enforcement if companies break their agreements. In fact, \textbf{the RSP framework allows companies to break their RSPs if they are concerned about risks from their competitors}. If a company believes that a competitor may cause unacceptably high risk, the company is allowed to break its RSP at its own discretion (Alignment Research Center, 2023; Anthropic, 2023). The RSP framework allows each company to determine what counts as an emergency and which competitors should be deemed sufficiently irresponsible. This underscores the need for approaches that have greater accountability and enforceability. 
    \item \textbf{The safeguards do not exist}. Experts openly acknowledge that they do not know how to control AGI. Anthropic’s RSP, for example, describes AGI-like systems as “ASL-4” systems. Anthropic openly acknowledges that it does not know what safeguards will be needed to control these systems, and it does not yet have a plan to ensure that these systems are developed safely (Anthropic, 2023). Anthropic is not unique in this regard – no other AI companies have yet developed safeguards that provably remove dangerous capabilities, nor produced plans that provably minimize extinction risk from Frontier AI. In the case where an evaluation successfully identifies a trained model as dangerous, there exist no countermeasures to neuter the threat. 
   \item \textbf{RSPs do not require concrete commitments. }Under RSPs, when a company realizes it has a dangerous system, it is not required to specify how it will proceed. For example, if a company successfully detects that its AI system can develop biological weapons or shows signs of escaping from human control, RSPs do not require companies to specify how they will respond. RSPs do not need to specify whether or not the company will pause further development or deployment, whether they will inform anyone about these risks, or how they will manage these risks. 
  \item \textbf{The tests are inadequate}. RSPs rely on \textit{dangerous capability evaluations} – tests that show whether or not a model is capable of performing dangerous activities in specific scenarios. This approach relies on having an exhaustive list of scenarios, an exhaustive list of dangerous capabilities, and an exhaustive set of evaluations that can track those capabilities with proven accuracy. None of these exist. Neither experts nor RSP proposals have concrete or comprehensive lists of dangerous scenarios, exhaustive lists of dangerous capabilities, or reliable evaluations for those capabilities. Looking for an exhaustive list of scenarios is a flawed exercise in itself, in a domain where capabilities of AI systems are regularly discovered long after they are deployed to millions of people. 
\end{enumerate}

\newpage
\section{The way forward: Policy recommendations}
\hfill\begin{minipage}{\dimexpr\textwidth-1cm}
“Many of the world’s leading AI experts now think that human-level (or even more capable) AI could arrive before 2030. Regardless of the exact timelines, it’s clear that unelected tech leaders should not decide whether, when, how and for what purpose such transformative AI is built.” – Bengio \& Privitera (2023)
\end{minipage}
\hspace{1cm}

In this section, we describe our preferred policy responses. We recommend the following three measures:

\begin{enumerate}
    \item \textbf{MAGIC: A multinational artificial general intelligence consortium}. Governments should establish a multinational institution to enforce a global compute cap, perform research on how to control highly powerful AI systems, develop emergency response infrastructure, and develop additional measures to reduce extinction risks from advanced AI (see Hausenloy et al., 2023). 
    \item \textbf{A global compute cap}. Governments should prohibit the development of AI systems above a predetermined amount of computing power. We recommend setting this cap on compute at 10\textsuperscript{24} floating point operations (FLOP), around the size of ChatGPT. If the cap cannot be implemented immediately, we should — at a minimum — develop the infrastructure and institutions necessary to implement such a cap in the future.
    \item \textbf{Gating critical experiments: }Governments should establish auditing regimes that demand \textbf{affirmative evidence of safety }for certain kinds of dangerous AI development. Companies developing AI systems above a certain compute threshold (but lower than the global compute cap) would be required to show affirmative evidence of safety. This regulatory regime could draw from practices in other high-risk sectors (e.g., nuclear safety). 
\end{enumerate}

\subsection{MAGIC: A Multinational Artificial General Intelligence Consortium}

\textbf{Governments should establish a Multinational Artificial General Intelligence Consortium }(\textbf{MAGIC; see Figure 1}). MAGIC would have an exclusive mandate to conduct critical Frontier AI research. This institution would adhere to national-security grade standards such as closed-off facilities, an isolated network, clearances, regular vulnerability testing, and strict access policies. Rather than developing advanced AI in the context of a race between corporations, MAGIC would be the world’s only artificial general intelligence project. It would be a highly secure organization that is tasked with understanding how to control highly advanced AI systems, ensuring that the benefits of safe AI systems are globally distributed, and preventing other actors from illegally developing advanced AI (Hausenloy et al., 2023). 

\begin{figure}[h]
    \centering
    \includegraphics[width=1\linewidth]{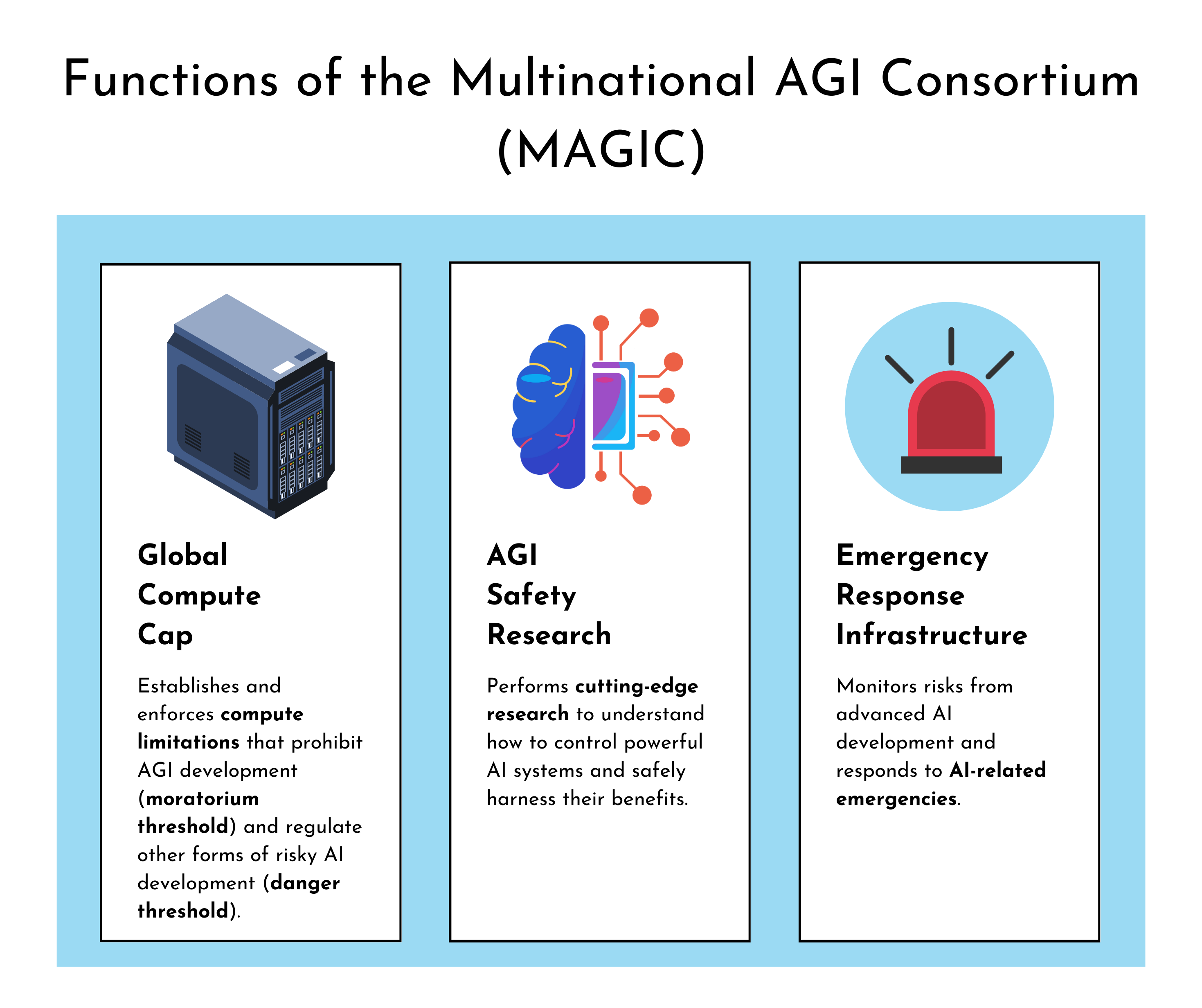}
    \caption{MAGIC would be responsible for implementing \textbf{compute limitations}, performing \textbf{AGI safety research}, and maintaining infrastructure to respond to \textbf{AI-related emergencies}.}
    \label{fig:enter-label}
\end{figure}

\textbf{MAGIC would allow the world to have democratic oversight over the development of highly powerful and highly dangerous AI systems}. MAGIC provides a positive and proactive vision forward: it enables innovation to occur once world leaders and AI experts have determined that such innovation would be safe and beneficial for humanity. 

\textbf{MAGIC would be the world’s unified AGI safety project. }MAGIC would hire talented researchers from around the world to perform research on how to control highly powerful AI systems and develop additional measures to reduce extinction risks from advanced AI. Thanks to the global compute cap (described below), MAGIC would be able to perform this research \textit{without} getting locked into a dangerous race to godlike AI. MAGIC could develop evaluations that provide affirmative evidence of safety, develop safeguards for increasingly powerful AI systems, and quantitatively estimate risks from advanced AI. Once MAGIC researchers and world leaders have compelling affirmative evidence that they can develop powerful AI systems safely, they would be enabled to do so.

\textbf{MAGIC would develop emergency response infrastructure that allows world leaders to respond to an AI emergency (see figure 2)}. This infrastructure would need to have at least three parts: \textbf{detection} (to make sure that government officials notice potential emergencies quickly), \textbf{alarms} (to ensure that those tracking AI risks can swiftly communicate risks to MAGIC and other relevant officials to ensure a swift and coordinated response), and an \textbf{emergency response} (a “kill switch” that allows regulators to swiftly halt AI development and deployment). 
The emergency response system would be tested periodically. This would involve practice drills, in which regulators simulate what would happen if national or international regulators determined that risks had exceeded an acceptable threshold. In these drills, regulators would implement their emergency response efforts to test their effectiveness in a real-world setting. For instance, they would determine how long it would take them to halt an AI training run or withdraw access to a deployed AI system. The drills would allow them to ensure that their emergency response system would work in the event of an actual emergency, ensure they have swift communication channels with Frontier AI developers and compute providers, and help them identify ways of improving the emergency response infrastructure. 

\begin{figure}[h]
    \centering
    \includegraphics[width=0.8\linewidth]{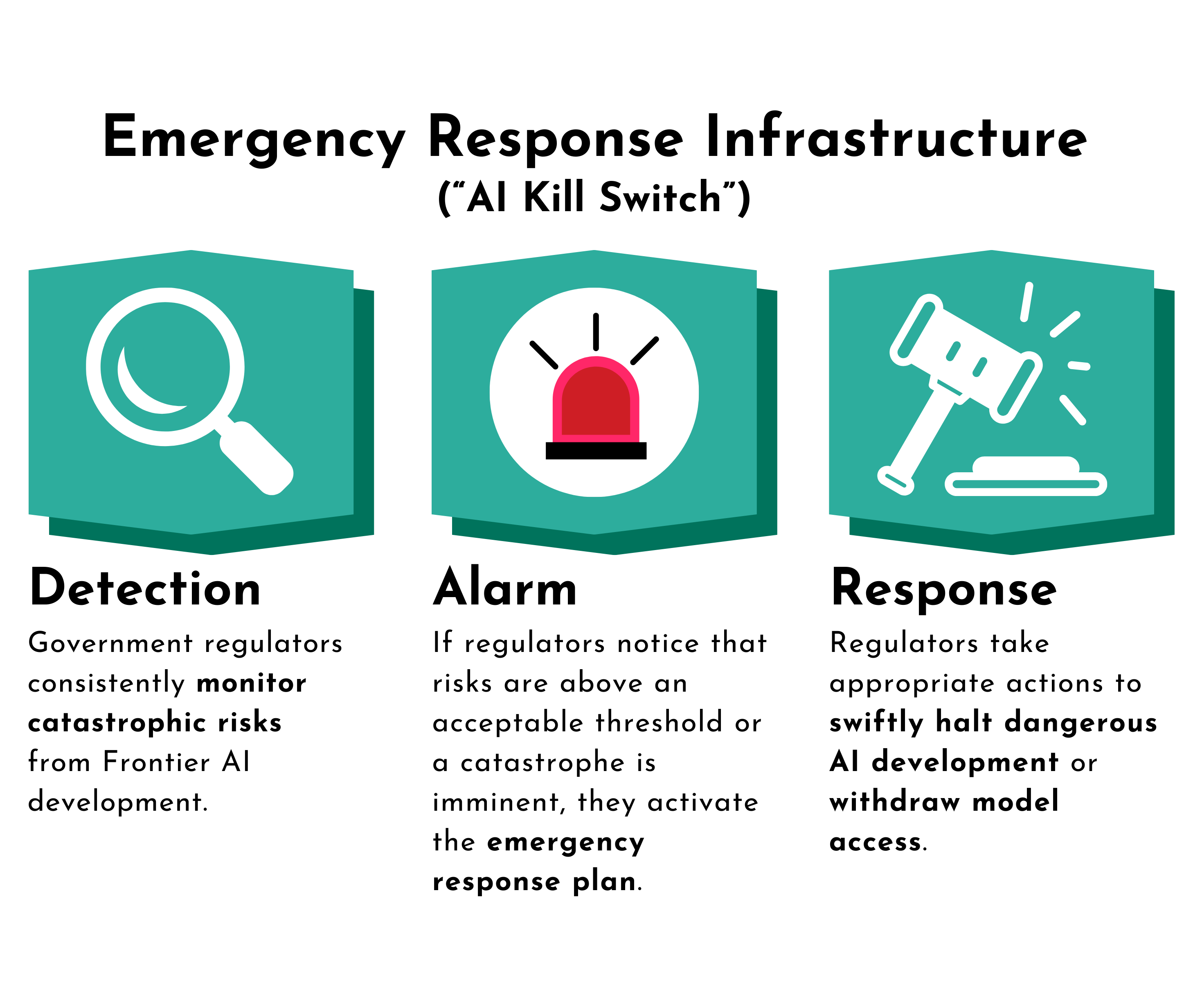}
    \caption{National and international regulators would develop infrastructure to \textbf{detect} elevated risk from AI, activate an \textbf{alarm} if risks exceed an acceptable threshold, and \textbf{respond} by halting Frontier AI development or withdrawing access to a deployed model.}
    \label{fig:enter-label}
\end{figure}

\subsection{Global Compute Cap}

\textbf{MAGIC would implement a tiered approach to compute limitations (see Figure 3)}. The most important compute limitation would be \textbf{a global moratorium on AI development above a certain amount of computing power}. Building state-of-the-art AI systems requires huge amounts of expensive machinery (called GPUs). This infrastructure is currently core to AI scaling, which makes it a bottleneck to truly dangerous systems. \textbf{Compute Limitations} address this bottleneck by enforcing a cap on the hardware used to build these AI systems. This mitigates extinction risk by throttling the very few dangerous AI systems that rely on advanced hardware, while leaving the rest of the AI ecosystem largely unhindered.

\begin{figure}[h]
    \centering
    \includegraphics[width=1\linewidth]{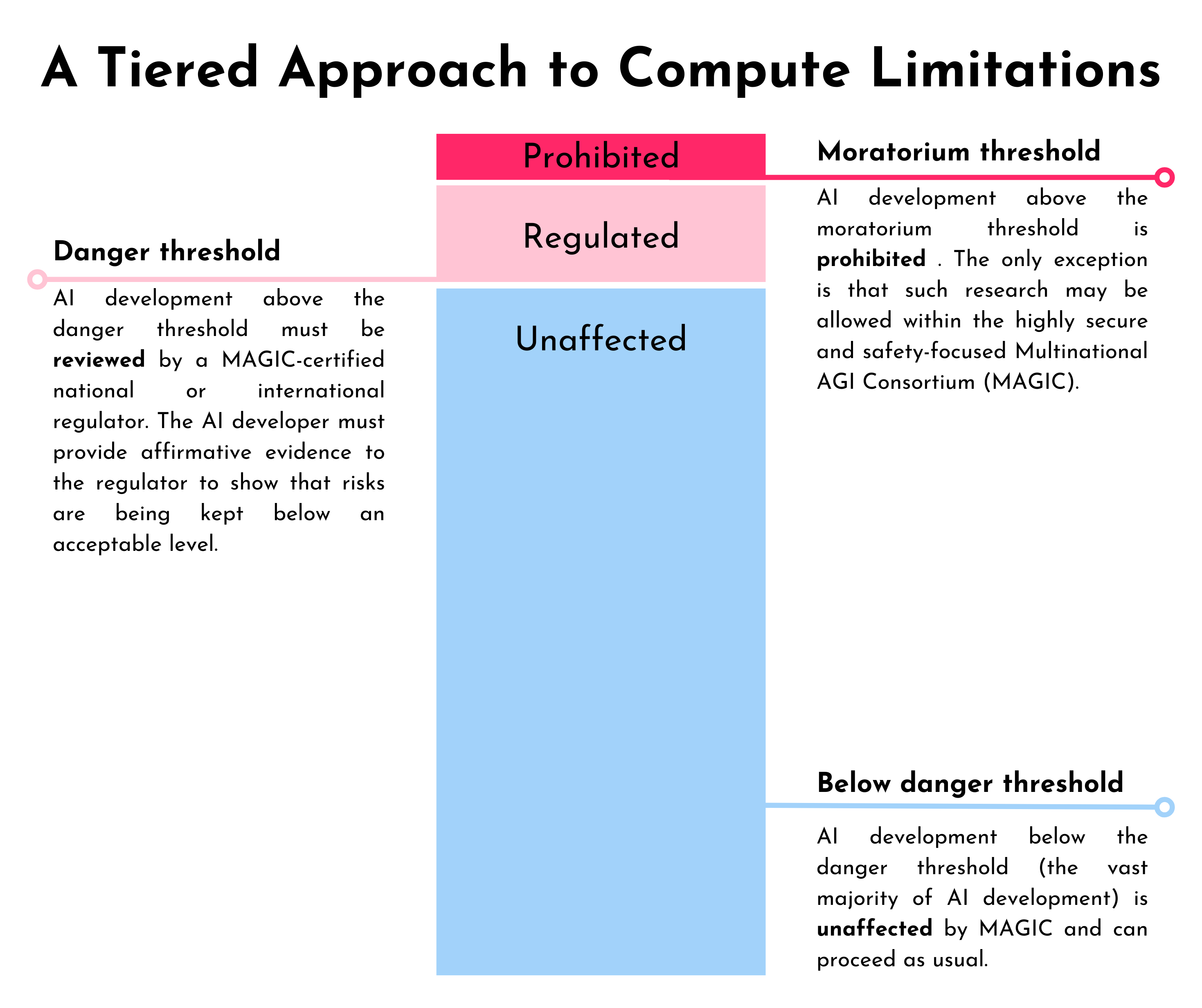}
    \caption{Under a tiered approach to compute limitations, AGI development would be forbidden above the \textbf{moratorium threshold}, regulated above a \textbf{danger threshold}, and \textbf{unaffected below} the danger threshold.}
    \label{fig:enter-label}
\end{figure}

Specifically, MAGIC would implement two compute limitations: \textbf{a moratorium threshold} (no one is allowed to develop AI systems above this threshold except MAGIC) and a \textbf{danger threshold} (companies are allowed to develop systems above this threshold but only if they can show affirmative evidence of safety, such that national or international regulators are convinced that extinction risks are acceptably low). AI development below the danger threshold would not be regulated by MAGIC, AI development above the moratorium threshold would be prohibited outside MAGIC, and AI development in between the thresholds would be governed by MAGIC-certified national regulators (see Figure 3). The danger threshold allows some kinds of risky AI development to occur in the commercial sector while still limiting extinction risks below a risk threshold (set by MAGIC). In the next section, we describe the kind of national regulations that would be needed for systems above the danger threshold.

\textbf{We have drafted an international treaty} that would establish a global moratorium and pave the way for an international AI governance organization (Miotti, 2023; see also Treaty on Artificial Intelligence Safety and Cooperation, 2023 for a related proposal).

\textbf{One of MAGIC's important roles will be to set and update the moratorium threshold}. If it is set too low, AGI could be developed outside MAGIC, considerably raising the chance of extinction and other AI-enabled catastrophes. If it is set too high, some safe AI development would be delayed. Furthermore, new algorithms and techniques are regularly discovered that can make AI systems more powerful at a given level of compute. Therefore, MAGIC will need to update these thresholds over time to account for new techniques and other developments in AI progress. We recommend setting the initial moratorium threshold at 10\^{}24 floating point operations (FLOP).

This threshold reflects a few important facts about the AI development landscape. First, AI performance has been greatly affected by the quality and quantity of advanced hardware. Moravec (1998) explained this trend in a seminal paper, in which he tracked growth in computing power, extrapolated future trends, and predicted when the processing power of computers would match the general intellectual performance of the human brain. His estimates suggested that we would reach AGI at some point in the 2020s (Moravec, 1998). Second, current AI systems already perform at or above human-level performance on a variety of intellectual tasks. For example, GPT-4 performs at 90th percentile on the Bar Exam, 88th percentile on the LSAT, and 99th percentile on the SAT verbal (OpenAI, 2023). Third, many contemporary AI experts believe AGI could be developed very soon. For example, Google DeepMind co-founder Shane Legg puts a 50\% chance of developing AGI by 2028, Anthropic CEO Dario Amodei believes human-level intelligence could emerge within 2-3 years (Patel, 2023a; Patel, 2023b). In summary, we already have AI systems that use human-level amounts of computing power, we already have AI systems that are at human-level or superhuman-levels on challenging cognitive tasks, AI experts believe that AGI could be developed very soon given current trends, and new algorithms and techniques are making AI systems more powerful (and dangerous) even at a given level of compute. If the threshold is set too low, AGI could be developed before the international community is confident that it can be developed and implemented safely. Meanwhile, the vast majority of AI development requires far fewer than 10\^{}24 FLOP. With all of this in mind, we believe an initial moratorium threshold of 10\^{}24 FLOP would be an appropriate starting point for the moratorium threshold. Ultimately, MAGIC will be responsible for deciding the initial threshold and how to update it over time.

\subsection{Gating critical experiments and affirmative safety evaluations }

\textbf{AI development above a certain threshold of computing power should be prohibited unless there is affirmative evidence of safety from AI developers}. Before building critical systems and deploying them to the public, companies should be required to demonstrate the safety of these systems through verifiable criteria. The results of these tests should be audited and verified by government inspectors. It should draw from well-established risk management principles and frameworks from other high-risk areas (e.g., nuclear safety, biosafety). 

\textbf{Gating critical experiments places the burden of proof on Frontier AI developers to show that their practices keep risks below an acceptable level}. This contrasts with the RSP approach, highlighted earlier. RSP frameworks that focus on dangerous capability evaluations reverse the burden of proof: they assume that dangerous AI scaling should be allowed to occur until risks are already realized. This is not an appropriate risk management approach, and it is not the approach we use to manage other high-risk technologies. Instead, we commonly expect individuals and companies that engage in dangerous experiments to provide affirmative evidence that their activities keep risks at an acceptable level (see Raz \& Hillson, 2005). If they cannot demonstrate affirmative evidence of safety, they are not allowed to run dangerous projects or experiments. An affirmative safety regulatory regime would require AI developers to show their systems are safe, well-understood, and controllable. At a bare minimum, governments should require compelling evidence that AI systems will not autonomously deceive humans. 

\textbf{Gating critical experiments could be informed by standard practices from high-risk sectors, where demonstrating safety is a precondition for undertaking high-risk endeavors.} For example, companies might have to show evidence that they understand how their AI system reaches conclusions, they can make sure models don’t access a certain fact or set of facts, they can prevent AI models from ever generating certain content, they are aware of all the facts the model knows, they have a convincing case that the AI system cannot autonomously deceive people, and they have evidence that the cost of jailbreaking their model is higher than a certain threshold (e.g., >\$100,000). Of course, these particular criteria are simply meant to be illustrative examples; in practice, AI experts do not know how to show affirmative safety. This is concerning. As a result, companies developing sufficiently dangerous may have to explore novel approaches in order to present sufficient evidence and show regulators that risks are kept below an acceptably low level.

Importantly, such methods could draw from \textbf{probabilistic risk assessments} in the nuclear domain. In nuclear safety, it is common to set a concrete risk level that cannot be surpassed. For example, the US Nuclear Regulatory Commission (NRC) sets an acceptable risk threshold of increased cancer rates at 0.1\%; those who want to develop nuclear reactors must show that the reactor has a <0.1\% chance of increasing cancer risks for residents living near the facility (Nuclear Regulatory Commission, 1986). 

\subsection{A timeline for regulation}

These proposals will require swift international coordination. Given that some experts are expecting highly dangerous AI systems within the next 2-5 years (e.g., Oversight of A.I.: Principles for Regulation, 2023a) there is a need for urgent action. 

We envision the following as one pathway toward international coordination (see also Aguirre, 2023). 

\begin{enumerate}
    \item \textbf{National compute cap}. One major country implements a national compute cap. For example, the Executive Branch of the United States or the United Kingdom issues an order prohibiting the development of AI systems above a certain amount of training FLOP. 
    \item \textbf{\textbf{National emergency response infrastructure}.} One major country develops “kill switch” infrastructure. All AI development above a FLOP threshold must be registered in a government database. The government establishes a direct line of communication with executives and researchers at any companies developing these systems. The government employs a small team of people to monitor risks from Frontier AI development and activate an emergency response protocol if risks are determined to be above a predetermined threshold. The emergency response protocol involves the government swiftly halting the development of an AI system or withdrawing the deployment of an AI system (by coordinating with the AI company and/or the compute provider). Periodically, the government runs a “drill” to see if it can swiftly halt the development of a Frontier AI system or withdraw access to a deployed AI system. 
    \item \textbf{\textbf{International coordination}.}
\begin{enumerate}
        \item The first country to implement this regime drafts an international treaty to establish a global moratorium and a multinational international AGI safety organization (e.g., Miotti, 2023).
        \item 6 months after the UK AI Safety Summit, another summit convenes in which countries are presented with the Treaty.
        \item 12 months after the UK AI Safety Summit, most of the major AI world powers have signed the treaty. Signatory countries must demonstrate compliance with the international organization. 
        \item At least once a year, signatories meet to discuss the implementation of the treaty, share information about the risks from advanced AI, and discuss possible changes to the global moratorium threshold.
\end{enumerate}

\end{enumerate}

Below, we offer additional details on each of these steps.

\textbf{National compute cap}. Immediately, we believe that the nations leading AI development – the United States and the United Kingdom – should take the lead on implementing a national compute cap and a national emergency response infrastructure. The most advanced and dangerous AI systems are currently developed by companies based in the United States, making the United States’s leadership on this issue particularly important. To successfully implement a national compute cap, a government would need to monitor advanced AI development. Fortunately, advanced AI development currently requires extremely large quantities of advanced hardware, and there are only a few major data centers that are powerful enough to support advanced AI development. Monitoring advanced AI development could draw from some techniques used to monitor nuclear development (Baker, 2023) and may eventually involve hardware-based monitoring techniques (Shavit, 2023). Importantly, the kind of hardware used for advanced AI development differs substantially from the kind of hardware used for consumer purposes. As a result, such monitoring could be implemented without unnecessary externalities to ordinary consumers. 

\textbf{Emergency response infrastructure}. For the emergency response infrastructure, governments can draw from existing best practices in risk monitoring and emergency preparedness. This infrastructure should be developed alongside national security experts and intelligence experts who already have experience developing and implementing emergency response plans. In the case of advanced AI development, an adequate emergency response system would have at least two components: (a) the ability to identify an imminent emergency and (b) the ability to swiftly respond. The first component involves a \textbf{risk awareness}\textbf{ and monitoring system}. This could involve direct monitoring of data centers with high concentrations of compute, a direct communication line between Frontier AI developers and government officials, whistleblower protections for individuals who report dangers, and government audits of AI facilities. If the government receives evidence of highly dangerous AI development (e.g., a system that can develop biological weapons or escape human control), the government would then need to have the ability to swiftly halt AI development or withdraw AI systems that are widely deployed. This involves a “\textbf{kill switch}”, in which a government could swiftly contact AI developers or compute providers and tell them to stop an AI training run or withdraw API access to a model. 

To ensure that this infrastructure would work in an actual emergency, the government should perform “drills” at regular intervals. That is, once every few months, the government should simulate a high-risk scenario, ensuring that it can successfully halt AI training runs and successfully withdraw deployment of AI systems. For example, a test could involve simulating a scenario in which a new prompting technique allows ChatGPT users to commit highly dangerous cyberattacks. To test the emergency response protocol, the government would swiftly get in touch with OpenAI and coordinate to withdraw ChatGPT for a few minutes. 

\textbf{International coordination}. A few months after the national regulations are implemented, the national regulations serve as a foundation for international coordination and the establishment of MAGIC. MAGIC would enforce a global compute cap and take other measures to reduce risks from advanced AI (Hausenloy et al., 2023). All major AI powers (e.g., the United States, the United Kingdom, and the People’s Republic of China) recognize the risks and agree to avoid a nationalistic race to advanced AI. MAGIC provides incentives for others to join (e.g., export controls on advanced hardware for states that do not join; access to technical expertise from MAGIC for states that do join). In addition, typical diplomatic means (e.g., sanctions) are employed to incentivize states to join, and typical diplomatic measures are taken to prevent non-signatories from developing smarter-than-human AI systems.

\subsection{Concrete recommendations for the UK AI Safety Summit}

Table 1 describes concrete commitments that governments could make, both immediately and in the longer term.

\begin{table}[hbt!]
\caption{Example international measures that minimize AI extinction risk}
\SetTblrInner{rowsep=5pt,colsep=7pt}
\begin{tblr}{|Q[3cm,valign=m,halign=l]|X[j,valign=m]|X[j,valign=m]|}
        \hline
        \textbf{Policy} & \textbf{Example international\\ statement}  & \textbf{Example action from world leaders} \\
        \hline 
        Multinational ‘CERN for AI’ (MAGIC) & “Most urgently, \textbf{we commit to the establishment of an international institution} with the exclusive mandate to conduct critical Frontier AI research. We recognize that this is a long-term solution to reach a stable, global equilibrium on AI that minimizes AI extinction risk. & The UK announces that it will be convening another summit in 3 months for the explicit purpose of establishing a multinational AGI safety organization (MAGIC). At the summit, world leaders present an international treaty to establish MAGIC, a global compute cap, and other safety provisions (see Miotti, 2023).   \\
        \hline
        Compute limitations (global compute cap) & “We affirm our commitment to mitigating extinction risk from AI by agreeing to \textbf{limit computing power available for large-scale, dangerous AI development}. This limit shall be revised over time to account for advances in the technology.” & The UK can lead on this commitment by setting an adequate threshold (such as 10\^{}24 FLOP), rallying a transatlantic alliance to move ahead with such measures, and setting the ground for more comprehensive international treaties. \\
        \hline
        Gating critical experiments (government-mandated affirmative safety evaluations) & “We recognize the risky nature of Frontier AI research, and thus endorse the principle that \textbf{the onus should be on companies to demonstrate the safety of their systems ahead of developing and deploying them, as is standard practice in other high-risk sectors”}. & The UK can pioneer the development of Safety Proofs by establishing a minimal list of properties that AI developers should demonstrate. The UK AI Foundation Model Taskforce could be in charge of ensuring that AI developers go through this process before each critical experiment and deployment to the public. \\
        \hline
\end{tblr}
\end{table}

\newpage
\newpage
\section{Conclusion}
Advanced AI poses extinction risks to humanity, voluntary commitments are not a sufficient response to this problem, and governments have an enormous opportunity to intervene. Ian Hogarth, chair of the UK’s AI Foundation Model Taskforce, recently expressed the need to slow the race to godlike AI to avoid an AI catastrophe (Hogarth, 2023). We strongly support this goal, and we believe it is a necessary and responsible step to reduce extinction risks. In this report, we described how “responsible scaling” and other forms of voluntary commitments will be insufficient, argued that affirmative evidence of safety should be required from advanced AI developers, and proposed measures that world leaders could take to substantially reduce risks from advanced AI. 

We look forward to following the outcomes of the UK AI Safety Summit, and we appreciate the opportunity to offer our input. Hopefully, the UK AI Safety Summit will provide a forum for world leaders to meaningfully discuss the extreme threats posed by advanced AI and begin to reach a consensus on the kinds of policy solutions that are needed. It will be important that those policy solutions go beyond voluntary commitments that allow AI companies to continue racing toward godlike AI. It will also be important for stronger solutions – such as global compute caps and emergency response infrastructure – to be treated as urgent priorities. 

The world may not have many years left before the development of highly dangerous AI systems. The Summit may represent one of the world’s key chances to curb extinction risks from advanced AI. We hope world leaders make use of this remarkable opportunity. 

\newpage
\section*{Appendix}\label{sec:intro}
\subsection*{Extinction risks from advanced AI} 
\textbf{The proposals we outlined have considerable support from the US and UK public}. The AI Policy Institute (AIPI) conducts polls of the American public (AIPI, 2023; Schreckinger, 2023). 

Their polls have found that:
\begin{itemize}
    \item 83\% believe AI could accidentally cause a \textbf{catastrophic event.}
    \item 82\% \textbf{prefer slowing down the development of AI} (compared to 8\% who prefer speeding it up).
    \item 82\%\textbf{ do not trust AI tech executives }to regulate AI.
    \item “Preventing dangerous and catastrophic outcomes” from AI was ranked as the most important policy priority.
\end{itemize}

Control AI commissioned a poll of the UK electorate (Control AI, 2023; YouGov, 2023). The poll found:
\begin{itemize}
    \item 78\% support the creation of a “\textbf{global watchdog}” to regulate powerful AI.
    \item 74\% believe the government should \textbf{prevent AI from “reaching superhuman capabilities”.}
    \item 72\% believe AI should be treated as an “incredibly powerful and dangerous technology”.
\end{itemize}

\pagebreak

\bibliographystyle{apalike}
\bibliography{example}
\sloppy

Aguirre, A. (2023). \textit{Close the Gates to an Inhuman Future: How and why we should choose to not develop superhuman general-purpose artificial intelligence.} \url{https://papers.ssrn.com/sol3/papers.cfm?abstract_id=4608505}

Alignment Research Center. (2023). \textit{Responsible Scaling Policies (RSPs).} \url{https://evals.alignment.org/blog/2023-09-26-rsp/}

Altman, S. (2015). \textit{Machine intelligence, part 1.} \url{https://blog.samaltman.com/machine-intelligence-part-1}

Anthropic. (2023). \textit{Anthropic’s Responsible Scaling Policy.} \url{https://www-files.anthropic.com/production/files/responsible-scaling-policy-1.0.pdf}

Artificial Intelligence Policy Institute. (2023). \textit{Homepage - AI Policy Institute.} \url{https://theaipi.org/}

Baker, M. (2023). \textit{Nuclear Arms Control Verification and Lessons for AI Treaties. }\url{https://arxiv.org/abs/2304.04123}

Bengio, Y., Hinton, G., Yao, A., Song, D., Abbeel, P., Harari, Y. N., Zhang, Y.-Q., Xue, L., Shalev-Shwartz, S., Hadfield, G., Clune, J., Maharaj, T., Hutter, F., Baydin, A. G., McIlrath, S., Gao, Q., Acharya, A., Krueger, D., Dragan, A., Torr, P., Russell, S., Kahnemann, D., Brauner, J., \& Mindermann, S. (2023). \textit{Managing AI Risks in an Era of Rapid Progress.} \url{https://managing-ai-risks.com/managing_ai_risks.pdf}

Bengio, Y. \& Privitera, D. (2023). \textit{How We Can Have AI Progress Without Sacrificing Safety or Democracy.} \url{https://time.com/collection/time100-voices/6325786/ai-progress-safety-democracy/}

Center for AI Safety. (2023). \textit{Statement on AI Risk.} \url{https://safe-ai.webflow.io/statement-on-ai-risk}

Control AI. (2023). \textit{Poll Reveals British Voters Overwhelmingly Favour Regulation.} \url{https://responsiblescaling.com/blog-detail/polling}

Hausenloy, J., Miotti, A. \& Dennis, C. (2023). \textit{Multinational AGI Consortium (MAGIC): A Proposal for International Coordination on AI.} \url{https://arxiv.org/abs/2310.09217}

Hogarth, I. (2023) \textit{We must slow down the race to God-like AI.} \url{https://www.ft.com/content/03895dc4-a3b7-481e-95cc-336a524f2ac2}

Leike, J. \& Sutskever, I. (2023). \textit{Introducing Superalignment. }\url{https://openai.com/blog/introducing-superalignment}

Miotti, A. (2023). \textit{An International Treaty to Implement a Global Compute Cap for Advanced Artificial Intelligence.} \url{https://papers.ssrn.com/sol3/papers.cfm?abstract_id=4617094}

MIT Technology Review. (2023). \textit{Video: Geoffrey Hinton talks about the “existential threat” of AI.} \url{https://www.technologyreview.com/2023/05/03/1072589/video-geoffrey-hinton-google-ai-risk-ethics/}

Moravec, H., (1998). \textit{When will computer hardware match the human brain?} \url{https://jetpress.org/volume1/moravec.pdf}

Nuclear Regulatory Commission. (1986). \textit{Safety Goals for the Operations of Nuclear Power Plants.} \url{https://www.nrc.gov/reading-rm/doc-collections/commission/policy/51fr30028.pdf}

OpenAI. (2023). \textit{GPT-4 Technical Report.} \url{https://arxiv.org/pdf/2303.08774.pdf}

\textit{Oversight of A.I.: Principles for Regulation: Hearing before the Judiciary Committee Subcommittee on Privacy, Technology, and the Law,} U.S. Senate, 118th Congr. (2023a). (testimony of Dario Amodei). \url{https://www.judiciary.senate.gov/imo/media/doc/2023-07-26_-_testimony_-_amodei.pdf}

\textit{Oversight of A.I.: Principles for Regulation: Hearing before the Judiciary Committee Subcommittee on Privacy, Technology, and the Law,} U.S. Senate, 118th Congr. (2023b). (testimony of Yoshua Bengio). \url{https://yoshuabengio.org/wp-content/uploads/2023/07/Written-Testimony-and-biography-of-Yoshua-Bengio_U.S.-Senate-Judiciary-Subcommittee-on-Privacy-Technology-and-the-Law_25_07_2023.pdf}

Patel, D. (Host). (2023a, August 8). Dario Amodei (Anthropic CEO) - Scaling, Alignment, \& AI Progress [Podcast episode]. In \textit{Dwarkesh Podcast}.  \url{https://www.dwarkeshpatel.com/p/dario-amodei}

Patel, D. (Host). (2023b, October 26). Shane Legg (DeepMind Founder) - 2028 AGI, New Architectures, Aligning Superhuman Models [Podcast episode]. In \textit{Dwarkesh Podcast}. \url{https://www.dwarkeshpatel.com/p/shane-legg}

Raz, T. \& Hillson, D. (2005). \textit{A Comparative Review of Risk Management Standards.} \url{https://www.researchgate.net/publication/228633220_A_Comparative_Review_of_Risk_Management_Standards}

Schreckinger, B. (2023). \textit{One think tank vs. ‘god-like’ AI.} \url{https://www.politico.com/newsletters/digital-future-daily/2023/08/15/one-think-tank-vs-god-like-ai-00111325}

Shavit, Y. (2023). \textit{What does it take to catch a Chinchilla? Verifying Rules on Large-Scale Neural Network Training via Compute Monitoring. }\url{https://arxiv.org/abs/2303.11341}

Tegmark, T. [@tegmark]. (2023, October 17). \textit{“Responsible scaling” sounds like it came from the same lobbying firm that coined “clean coal”} [Tweet]. X. \url{https://twitter.com/tegmark/status/1714244697726599211}

The Logan Bartlett Show. (2023). \textit{Anthropic CEO on Leaving OpenAI and Predictions for Future of AI} [Video]. YouTube. \url{https://www.youtube.com/watch?v=gAaCqj6j5sQ&t=5885s}

\textit{Treaty on Artificial Intelligence Safety and Cooperation} [Treaty Blueprint]. (2023). \url{https://taisc.org/taisc}

YouGov. (2023). \textit{Control AI Survey Results}. \url{ https://d3nkl3psvxxpe9.cloudfront.net/documents/ControlAI_AI_231019.pdf}

\end{document}